\documentclass[12pt]{article}

\usepackage[margin = 1in]{geometry}

\newcommand{\argmax}{\mathop{\rm argmax}}

\usepackage{xcolor}

\usepackage{amsopn}

\usepackage{lmodern}
\usepackage{amsmath}

\usepackage{graphicx}

\usepackage{amssymb,amsmath}

\usepackage{relsize}

\usepackage{upgreek}

\usepackage{subfigure}

\usepackage{url}

\usepackage{graphicx}

\usepackage{algorithmic}
\usepackage{algorithm2e}

\def\one{\mathbf{1}}

\pdfoutput=1
\usepackage{amsmath, amssymb}
\usepackage{graphicx,psfrag,epsf}
\usepackage{enumerate}
\usepackage[numbers]{natbib}
\usepackage{url} % not crucial - just used below for the URL 

\DeclareMathOperator{\argmin}{argmin}

\def\finit{\hat{\mathbf{F}}^*_0}
\def\ffinal{\hat{\mathbf{F}}^*_{\infty}}
\def\fstar{f^*}
\def\sample{\{\mathbf{x}_i\}_{i\in [n]}}
\def\X{\mathcal{X}}
\def\pik{\Pi_{K}}

\def\R{\mathbb{R}}

\def\I{\mathbf{I}}
\def\x{\mathbf{x}}

\def\W{\mathbf{W}}

\def\w{\mathbf{w}}

\begin{document}

\def\spacingset#1{\renewcommand{\baselinestretch}%
{#1}\small\normalsize} \spacingset{1}

%%%%%%%%%%%%%%%%%%%%%%%%%%%%%%%%%%%%%%%%%%%%%%%%%%%%%%%%%%%%%%%%%%%%%%%%%%%%%%

  \title{\bf Clustering by Non-parametric Smoothing}
  \author{David Hofmeyr
  \hspace{.2cm}\\
    School of Mathematical Sciences, Lancaster University}
  \maketitle

\bigskip
\begin{abstract}

A novel formulation of the clustering problem is introduced in which the task is expressed as an estimation problem, where the object to be estimated is a function which maps a point to its distribution of cluster membership. Unlike existing approaches which implicitly estimate such a function, like Gaussian Mixture Models (GMMs), the proposed approach bypasses any explicit modelling assumptions and exploits the flexible estimation potential of non-parametric smoothing.
%
%Furthermore, because the estimation problem resembles a standard statistical estimation problem, the formulation easily admits the use of ensemble models. This fact is explored in the form of bootstrap aggregation and in clustering bigger data sets than is usually possible with flexible estimators.
%
%An intuitive approach for  natural 
%The proposed approach naturally selects an appropriate number of clusters to extract from a data set and is robust to its tuning parameters, while also providing high clustering accuracy.
An intuitive approach for selecting the tuning parameters governing estimation is provided, which allows the proposed method to automatically determine both an appropriate level of flexibility and also the number of clusters to extract from a given data set.
Experiments on a large collection of publicly available data sets are used to document the strong performance of the proposed approach, in comparison with relevant benchmarks from the literature. {\tt R} code to implement the proposed approach is available from \url{https://github.com/DavidHofmeyr/CNS}.
\end{abstract}

\noindent%

%\newpage
\spacingset{1.45} % DON'T change the 

\section{Introduction}
\label{sec:intro}

Cluster analysis refers to the task of partitioning a set of data into groups (or clusters) in such a way that points within the same cluster tend to be more similar than points in different clusters. This is not a well defined problem, and different interpretations of the clustering objective can lead to vastly different methods for identifying clusters. Some popular formulations of the clustering problem include (i) centriod based clustering, in which clusters are determined based on how data group around their central points~\citep{Leisch2006}; (ii) density based clustering, in which clusters may be seen as data dense regions which are separated from other clusters by regions of data sparsity~\citep{campello2020density}; (iii) graph based clustering, in which clusters are aligned with highly connected subgraphs, which are weakly connected to other such subgraphs, and where these clusters are typically identified using spectral graph theory~\citep{Luxburg2007}; and (iv) model based clustering, in which the data distribution is modelled as a mixture of parametric, typically Gaussian, components with each component representing a cluster~\citep{fraley2002model}.

In this paper we consider a novel formulation of the clustering problem, and frame it as an estimation problem where the object of interest is a function from the input space to the distributions of cluster membership. In the model based clustering framework such a function is induced directly by the model itself, however in our framework we bypass any assumptions on this function (and hence on the forms of the clusters), except that it is continuous. This allows us to estimate the function in a fully data driven way, using the principles of non-parametric smoothing. Our approach shares some similarities with both spectral clustering~\citep{Luxburg2007} and Markov chain clustering~\citep{MCL}, however our approach is fundamentally distinct both practically and philosophically. We give a detailed description of our problem formulation, as well as our approach for conducting estimation and clustering, in the following section. We also describe an intuitive data driven approach for automatically selecting both the level of flexibility in the non-parametric estimation and the number of clusters to extract. We then go on to report on results from practical experiments with the proposed approach in Section~\ref{sec:experiments}, before giving some concluding remarks in Section~\ref{sec:conclusions}.

\section{Implicitly Estimating a Clustering Function}\label{sec:method}

In this section we introduce our formulation of the clustering problem, and describe our method for estimation. We consider the natural statistical setting in which our observations, say $\sample$, where $[n] = \{1, ..., n\}$, arose independently from a distribution function $F_X$, with support $\X \subset \R^d$, which admits a density function $f_X$. Existing formulations of the clustering problem which adopt this setting include primarily
\begin{enumerate}
    \item Model-based clustering~\citep{fraley2002model}, in which $f_X$ is assumed to be a mixture density, i.e., $f_X = \sum_{k=1}^K \pi_k f_k$, where $\{\pi_k\}_{k \in [K]}$ are the \textit{mixture weights} and $\{f_k\}_{k \in [K]}$ are the component densities. Here clusters are defined through the function $C(\x) = \argmax_k \pi_k f_k(\x)$. Almost exclusively in practice the components are assumed to have a simple parametric form, with the most common being to model each with a Gaussian density.
    \item Density-based clustering~\citep{campello2020density}, in which clusters are defined as the components (maximal connected subsets) of a chosen \textit{level set} of the density. The level set of $f_X$ at level $\lambda \geq 0$ is defined as $\{\x \in \X|f_X(\x) \geq \lambda\}$, i.e. the set of points with density at least $\lambda$. In practice $f_X$ is typically estimated with a flexible non-parametric estimator, and so clusters manifest as regions of high data density which are separated from one another by regions of relative sparsity.
    \item Mean-shift~\citep{ComaniciuM2002} and related methods, in which clusters are defined as the \textit{basins of attraction} of the modes of $f_X$. The basin of attraction of a mode of $f_X$, say $\mathbf{m} \in \X$, is defined as the set of points for which an uphill gradient-flow (gradient ascent with infinitesimal stepsize) converges to $\mathbf{m}$. The name \textit{mean shift} arises from the fact that repeatedly ``shifting'' a point to the local average of the observations around it performs a gradient ascent on a non-parametric estimate of $f_X$ with an implicitly well controlled step-size, and so can be seen as approximating this gradient-flow.
\end{enumerate}

All three approaches are highly principled, but each has its limitations. Model based clustering limits the flexibility of the individual clusters through the parametric form of the component densities; density-based clustering requires selection of $\lambda$, and a decision needs to be made how to treat the points which do not fall in the chosen level set (this is not to mention the non-trivial problems of appropriately estimating the density and subsequently identifying the components of its level sets); and mean-shift clustering is typically computationally demanding to execute and has been found to perform poorly on relatively high dimensional data.

Our formulation is philosophically distinct from these, and makes only the assumption that there exists a continuous function $f^*:\X \to \pik$, where $\pik$ is the $K$ dimensional probability simplex, i.e., the collection of all probability mass functions on the set $[K]$, which appropriately reflects cluster membership probabilities over $\X$. That is, for $\x \in \X$ the quantity $f^*(\x)$ is the vector with $k$-th entry equal to the probability that $\x$ is associated with the $k$-th cluster. Continuity is a natural assumption for such a function, since it implies that points which are near to one another have similar probabilities of cluster membership. Mixture densities induce such a function, where if $f_X = \sum_{k=1}^K \pi_k f_k$ then we simply have $\fstar(\x) = \frac{1}{f_X(\x)}\left(\pi_1 f_1(\x), ..., \pi_K f_K(\x)\right)$, and so our formulation may be seen as a generalisation of this framework.

\subsection{Estimating $\fstar$}

A highly principled approach for estimating continuous functions is through non-parametric smoothing. The basic idea upon which non-parametric smoothing techniques are based is that of a ``local average'', wherein estimation of a continuous function, say $g$, at a query point $\x \in \X$, is determined by
\begin{align}\label{eq:nps}
    \hat g(\x) = \sum_{i=1}^n w_i(\x) \tilde g(\x_i),
\end{align}
where $\tilde g(\x_i)$ may be seen as a ``noisy'' observation of $g$ at $\x_i$ %\footnote{Arguably the most common context in which this principle is applied is that of regression, where observations $\{y_i\}_{i\in [n]}$ are assumed to be equal to $\{g(\x_i) + \epsilon_i\}_{i\in [n]}$, for some zero mean \textit{residual} terms $\{\epsilon_i\}_{i\in [n]}$.}
and $\mathbf{w}:\X\to \Pi_n:\x \mapsto (w_1(\x), ..., w_n(\x))$ is a weight function which produces a probability distribution over the observations $\sample$, and which concentrates its probability mass on those observations nearest the argument, $\x$. For example, in the popular $k$-Nearest-Neighbour ($k$NN) based estimation the weight function is given by $w_i(\x) = \frac{1}{k}$ for $i \in \mathcal{N}_k(\x)$, where $\mathcal{N}_k(\x)$ are the indices of the $k$ nearest points to $\x$ from among $\sample$, and $w_i(\x) = 0$ otherwise. %Concentration of the distribution $\w(\x)$ on the observations nearest $\x$ ensures $\hat g(\x)$ is primarily influenced by

The most common application of Eq.~(\ref{eq:nps}) arises in the regression context, where $\{\tilde g(\x_i)\}_{i\in [n]}$ are the \textit{response variables}, typically denoted $\{y_i\}_{i\in [n]}$, and assumed to be equal to $\{g(\x_i) + \epsilon_i\}_{i \in [n]}$, for some zero mean \textit{residual} terms $\{\epsilon_i\}_{i \in [n]}$. Although in our context we do not have direct access to noisy observations of $\fstar$, %it is worth pointing out the potential of applying this approach when a prior estimate for $\fstar$ at $\sample$ is available, perhaps from applying a simpler clustering method.
%
%Repeatedly applying 
%
%In our context we do not assume access to an
%
%initially have noisy observations of the function $f^*$ on which to apply local averaging.
\textit{repeatedly} applying non-parametric smoothing to even very coarse approximations, in order to incrementally shift the estimates towards appropriate values, can be very effective. In fact we have found that even for completely random initialisations there is almost always a number of iterations of this approach which leads to highly accurate clustering solutions. Moreover, when data driven initialisations are used the quality of clustering solutions obtained can be extremely high. However, there are obvious limitations with such an approach. An appropriate number of iterations cannot be known \textit{a priori}, as the appropriateness of different numbers of iterations will clearly depend on the data and the weight function being used. 
%
%%%%%%%%%%%%%%%%%%%%%%%%%%%%%%%%%%%%%%%%%%%%%%%%%%%%%%%%%%%%%%%%%%%%%%%%%%%%%%%%%%%%%%%%%%%%%%%%
\iffalse

\begin{figure}[]
    \centering
    %\subfigure[Initialisation]{\includegraphics[width=0.48\linewidth]{knnclust1.pdf}}
    \subfigure[10 Iterations]{\includegraphics[width=0.48\linewidth]{knnclust2.pdf}\label{fig:iterative1}}
    \subfigure[20 Iterations]{\includegraphics[width=0.48\linewidth]{knnclust3.pdf}\label{fig:iterative2}}
    \subfigure[100 Iterations]{\includegraphics[width=0.48\linewidth]{knnclust5.pdf}\label{fig:iterative3}}
    \subfigure[1 000 Iterations]{\includegraphics[width=0.48\linewidth]{knnclust6.pdf}\label{fig:iterative4}}
    \subfigure[10 000 Iterations]{\includegraphics[width=0.48\linewidth]{knnclust7.pdf}\label{fig:iterative5}}
    \subfigure[100 000 Iterations]{\includegraphics[width=0.48\linewidth]{knnclust8.pdf}\label{fig:iterative6}}
    \caption{Iterative smoothing applied to a sample from a fifteen component Gaussian mixture with high overlap. Despite completely random initialisation the solutions after 100 and 1 000 iterations provide a fairly accurate clustering solution.}
    \label{fig:iterative}
\end{figure}

\fi
%%%%%%%%%%%%%%%%%%%%%%%%%%%%%%%%%%%%%%%%%%%%%%%%%%%%%%%%%%%%%%%%%%%%%%%%%%%%%%%%%%%%%%%%%%%%%%%%
%
%An appropriate number of iterations cannot be known \textit{a priori}, however, and in addition the appropriatness of different numbers of iterations will depend heavily on the data and the weight function being applied. 
More importantly, applying such an approach \textit{ad infinitum} will frequently yield convergence to solutions which are meaningless from the point of view of clustering. This is because iterative smoothing using the same weight function may be seen as formulating a Markov chain over the observations, for which the transitions out of observation $\x_i$ are determined by the distribution $\w(\x_i)$. The limiting behaviour of this iterative smoothing is therefore strongly dictated by the \textit{components} of this Markov chain, where two points are in a component if and only if they are mutually reachable from one another (after finitely many transitions of the chain), and each is reachable from every point which is reachable from it (i.e., they are \textit{essential}). In particular if every point is reachable from every other point then typically this iterative smoothing will converge to a constant function (in which all points have the same probabilities of cluster membership).

Clearly allowing convergence to a constant function is undesirable, and moreover clustering directly according to the components of the chain, even if there are more than one, is not robust to potential noise leading to connections which merge otherwise naturally separated clusters. The Markov chain analogy is, however, intuitively pleasing, since large values in its \textit{transition matrix} $\W$ (with $i$-th row equal to $\w(\x_i)$), which are interpreted as high probability transitions, are associated with pairs of points which are near to one another. We can imagine the behaviour of such a chain as moving freely and frequently between nearby points, or within clusters of nearby points, but only occasionally transitioning to points which are further away (and more likely to be in different clusters). Modelling clusters as collections of points within which the chain tends to spend a lot of time before departing is natural, and can be achieved through the very popular spectral clustering~\citep{Luxburg2007}. Intuitively appealing, and showing very strong performance in numerous practical applications, this formulation nonetheless suffers from being ill-posed in that ``a lot of time'' is not well-defined. 
%, and suffers similarly to the approach of terminating an iterative method prior to convergence.
Indeed, the spectral clustering methodology only produces a ``soft'' solution, and a final (separate) clustering approach is required to actually produce an output. This is not to mention the fact that determining the number of clusters when using spectral clustering is non-trivial except in a perfectly noise free scenario.\\
\\
Describing clusters through the limiting behaviour of the chain, rather than its finite-term behaviour, is clearer cut and avoids commitment to what is meant by ``a lot of time''. However, as mentioned previously the limiting properties of the chain may be meaningless for clustering, and are at best very sensitive to noise. Existing approaches which are motivated by this limiting behaviour typically rely on what we believe are inelegant tricks to break the natural convergence of the chain. For example, the Markov Clustering algorithm~\citep[MCL]{MCL} modifies the iterative updates to the probabilities of cluster membership to include so-called \textit{inflation} and \textit{expansion} steps which both ``inflate'' the larger probabilities and ``deflate'' the smaller ones. This inevitably introduces multiple tuning parameters, which are not always intuitive, in addition to those needed to define the weight function. Moreover these approaches often need considerable compute time to implement, as their algorithms explicitly apply recursive updating rather than the well established theoretical properties of Markov chains. %As a further consequence, a threshold to determine numerical convergence must be determined, selection of which is non-trivial in this context.

We take a fundamentally different approach, which simply modifies the iterative smoothing formulation to include a very small weight for the initial solution. That is, for $\lambda \in (0, 1)$, and starting with an initial solution $\{\hat f^*_0 (\x_i)\}_{i \in [n]}$, our approach is based on the update
\begin{align}\label{eq:update}
    \hat f^*_{t+1}(\x_i) &=
        (1-\lambda)\sum_{j=1}^n w_j(\x_i)\hat f^*_t(\x_j) + \lambda \hat f^*_0(\x_i).
\end{align}
Within the Markov chain analogy this is equivalent to introducing $K$ additional states (one for each potential cluster), which are \textit{absorbing}. In addition to transitions between pairs of observations, as in the chain described previously, at every time point there is a fixed probability, equal to $\lambda$, that the chain enters one of the absorbing states and then never leaves. Which absorbing state is entered, when transitioning from point $\x_j$, is determined by the distribution $\hat f^*_0(\x_j)$. With this interpretation the solution to which Eq.~(\ref{eq:update}) converges is simply the vector of probabilities that the chain, if starting in $\x_i$, eventually lands in each of the absorbing states. 

We find this formulation far more natural than the modifications in existing methods, and very importantly this formulation also admits a closed form solution,
\begin{align*}
    \lim_{t\to\infty} \hat f^*_t(\x_i) = \lambda \left(\mathbf{I} - (1-\lambda) \W\right)_{i:}^{-1}\hat{\mathbf{F}}^*_0,
\end{align*}
where $\left(\mathbf{I} - (1-\lambda) \W\right)_{i:}^{-1}$ is the $i$-th row of $\left(\mathbf{I} - (1-\lambda) \W\right)^{-1}$; $\mathbf{W}$ is the transition matrix from the original Markov chain (with $i$-th row $\w(\x_i)$) and $\hat{\mathbf{F}}^*_0$ is the matrix with the initial solutions, $\{\hat f^*_0(\x_i)\}_{i \in [n]}$, stored row-wise. This convergence relies on elementary Markov chain theory, where the modified chain (including the absorbing states) has transition matrix
\begin{align*}
    \tilde \W = \left[\begin{array}{cc}
        (1-\lambda) \W & \lambda \hat{\mathbf{F}}^*_0 \\
        \mathbf{0} & \I
    \end{array}\right].
\end{align*}
Compared with the iterative approach the value of $\lambda$ acts roughly inversely to the number of iterations, however we have found selecting $\lambda$ is far easier than selecting the number of iterations and, as mentioned above, this formulation admits a closed form solution.

\subsection{Practicalities and Tuning}

In this subsection we describe some of the practicalities related to the implementation of our approach, and propose a fully data driven criterion for selecting all of its tuning parameters. This is extremely important since appropriately selecting the level of flexibility in a non-parametric estimator is notoriously challenging, and validation of clustering models is typically not realistic unless some domain knowledge is available.

%Although we have found this approach to yield quite consistently good performance, we nonetheless recommend some investigation of the results in practice, especially when domain knowledge is available which may be used to validate any solutions obtained.

\subsubsection{The Weight function and setting of $\lambda$}

There are multiple popular approaches for determining the weights in a non-parametric smoothing estimator, such as nearest neighbours and kernels. Any of these can render accurate estimation, provided an appropriate choice of their smoothing parameters is made. Generally speaking, estimation is more flexible for smaller values of these smoothing parameters, but the added flexibility comes at the cost of greater sensitivity to noise. We use nearest neighbour weights as they are advantaged over many others in terms of computational speed. This is perhaps especially pronounced in the proposed approach, where computing the final solution, in the rows of $\ffinal := \lambda (\I - (1-\lambda)\W)^{-1}\finit$, can only be performed in a reasonable amount of time on large data sets if $\W$ is sparse.

The effect which the parameter $\lambda$ has on the solution to the proposed formulation may be easily understood, since it directly controls the extent to which the initial solution, in the rows of $\hat{\mathbf{F}}^*_0$, is allowed to influence the final clustering. The appropriateness of different choices for $\lambda$ therefore also differ depending on $\finit$. There is also a level of interplay between the value of $\lambda$ and the smoothing parameter, which in our case is the number of neighbours~($k$).

%a simple heuristic for automatically setting the tuning parameters for the proposed approach, which we have found to provide quite reliably good performance.

%Broadly speaking we have found values for $\lambda \in [0.001, 0.01]$ to be most reliable, and if $\finit$ is chosen so as not to strongly indicate a clustering solution (e.g., with almost all rows being uniform) then setting $\lambda = 0.01$ is a particularly reliable setting. For the number of neighbours in the weight function we simply use a single sample size dependent setting of $k = \lfloor \min\{0.5\sqrt{n}, 2\log(n)\}\rfloor$.

\subsubsection{Initialisation}

Aligning with the adage of letting the data speak for themselves, we employ initialisations which do not strongly indicate a clustering solution, so that the flexible non-parametric smoothing is able to guide estimation appropriately. However, it is important to note that if $\finit$ is completely uniform then so too will be the final solution. In other words $\finit$ must include \textit{some} information to differentiate points in potential clusters, to then be propagated by the weight function. How we achieve this is to, for a given value of $K$, find a collection of $K$ points, say $\{\x_{i^*}\}_{i \in [K]}$, which are likely to mostly belong to distinct clusters. We then set each row of $\finit$ to be uniform on $[K]$, except for the rows corresponding to observations $\{\x_{i^*}\}_{i \in [K]}$, which are set respectively to each of the $K$ indicator vectors. In other words, the points $\{\x_{i^*}\}_{i \in [K]}$ are initially given their own clusters and all other points are initially equally likely to be in each potential cluster. %To find $\{\x_{i^*}\}_{i\in [K]}$ we simply choose the points nearest the centroids from a $K$-means solution.

This approach, of having $\finit$ almost uniform, is pleasing for what it represents in terms of estimation, but also has the computational advantage that the final solution only depends on the columns of $\lambda(\I-(1-\lambda)\W)^{-1}$ associated with the indices $\{i^*\}_{i \in [K]}$. The reason for this is that we may write
\begin{align*}
    \finit =  \frac{1}{K}\one_n\one_K^\top + \mathbf{E}(\{i^*\}_{i\in [K]})
    -\frac{1}{K} \left(\sum_{j=1}^K \mathbf{E}(\{i^*\}_{i\in [K]})_{:j}\right)\one_K^\top,
\end{align*}
where $\one_{\cdot}$ is a vector of ones of given length; $\mathbf{E}(\{i^*\}_{i \in [K]}) \in \R^{n\times K}$ is a matrix of zeroes except in positions $(i^*,i); i \in [K],$ where it takes the value one; and the subscript ``$:j$'' indicates the $j$-th column of the matrix. Then, since $\lambda(\I-(1-\lambda)\W)^{-1}\one_n = \one_n$, we find that the final solution can be expressed as
\begin{align*}
    \ffinal =  \frac{1}{K}\one_n\one_K^\top + \lambda\left[\left(\I - (1-\lambda)\W\right)^{-1}_{:j^*}\right]_{j\in[K]}
    - \frac{\lambda}{K}\left(\sum_{j=1}^K \left(\I - (1-\lambda)\W\right)^{-1}_{:j^*}\right) \one_K^\top,
\end{align*}
where $\left[\left(\I - (1-\lambda)\W\right)^{-1}_{:j^*}\right]_{j\in[K]} \in \R^{n\times K}$ is the matrix with $j$-th column $\left(\I - (1-\lambda)\W\right)^{-1}_{:j^*}$.

We exploit this computational advantage in how we select $\{\x_{i^*}\}_{i\in [K]}$, since it allows us to choose these points from a larger candidate set based on their perceived prominence (we want to choose points which create clusters of non-negligible probability mass) and uniqueness (we don't want to choose multiple points which essentially represent the same clusters) within any \textit{final} solutions which contain them. That is, we can compute the columns of $(\I-(1-\lambda)\W)^{-1}$ for a set of indices $\{i'\}_{i \in [K']}; K' \geq K$, and use these to select a subset $\{i^*\}_{i \in [K]}$ as our final collection, based on the magnitude and uniqueness of these columns. That is, we define
\begin{align*}
    s_j &= \left\|(\I - (1-\lambda)\W)^{-1}_{:j}\right\|_1; j \in [n],\\
    c_{j,l} &= \left((\I - (1-\lambda)\W)^{-1}_{:j}\right)^\top(\I - (1-\lambda)\W)^{-1}_{:l}; j \not = l \in [n],\\
    c_{j,j} &= \infty; j \in [n],
\end{align*}
and then set
\begin{align*}
    1^* &= \argmax_{j \in \{1', ..., K''\}} s_{j},\\
    i^* &= \argmin_{j \in \{1', ..., K''\}} \max_{l \in \{1^*, ..., (i-1)^*\}} c_{j,l}/s_j^2; i \in \{2, ..., K\}.
\end{align*}
Specifically, the first point, $\x_{1^*}$, is the element of $\{\x_{i'}\}_{i\in [K']}$ whose corresponding column in $(\I-(1-\lambda)\W)^{-1}$ has the greatest magnitude, and subsequent points, $\x_{i^*}; i \in \{2, ..., K\}$, are chosen both to have columns with large magnitude and small inner products, hence similarities, with other columns already selected. The setting of $c_{j,j} = \infty$ is merely a convenience to avoid re-selection of the same indices into $\{i^*\}_{i \in [K]}$. Also, although we define the quantities $s_j, c_{j,l}$ for all $j, l \in [n]$, we only need to compute these values (and the associated columns of $(\I-(1-\lambda)\W)^{-1}$) for indices $\{i'\}_{i \in [K']}$. Following the objective of selecting points which would likely lead to prominent and unique clusters in any corresponding final solutions, we choose the indices $\{i'\}_{i\in [K']}$ to be the ``local'' maxima in the magnitudes of the columns of $\W$. Specifically, we add index $i \in [n]$ to this set iff $||\W_{:i}||_1 \geq \max_{j \in \mathcal{N}_k(\x_i)}||\W_{:j}||_1$, where, again, $\mathcal{N}_k(\x_i)$ is the set of indices of the $k$ nearest neighbours of $\x_i$.

\subsubsection{Automatic Tuning}

Here we describe a single criterion for the proposed approach which allows us to automatically determine appropriate settings for all of its tuning parameters, i.e., of $\lambda, k$ and $K$. The principle on which it is based is that if the settings are appropriate then the iterative smoothing should be able to substantially improve on the initial solution, where this initial solution, as described previously, is designed to encode minimal clustering information. We quantify this via the improvement in ``clarity'' of the clustering assignment,
\begin{align}\label{eq:CK}
    \nonumber
    C(\lambda, k, K) &:= \frac{1}{n}\sum_{i=1}^n\left(\max_{j\in[K]} (\ffinal)_{i,j} - \max_{j\in[K]}(\finit)_{i,j}\right)\\
    &= \frac{1}{n}\sum_{i=1}^n\max_{j\in[K]} (\ffinal)_{i,j} - \frac{n-K+K^2}{nK}.
\end{align}
Now, although $C(\lambda, k, K)$ depends implicitly on $\lambda$ and $k$ through their effect on the values in $\ffinal$, the reference value against which the clarity in the final solution is compared, i.e. the second term in Eq.~(\ref{eq:CK}) above, depends only on $K$. To incorporate the effects of $k$ and $\lambda$, we normalise $C(\lambda, k, K)$ by a second reference value which arises under an idealised scenario of ``perfect clusterability''. Specifically, in the idealised scenario (where the clustering information in $\finit$ is propagated optimally) we would have every observation in each cluster with the ``informative member'' of the cluster, i.e. the corresponding element of $\{\x_{i^*}\}_{i \in [K]}$, as one of its $k$ neighbours. Since at initialisation all ``non-informative points'' have the same probabilities of cluster membership, we would have for each $j \in [K]$ and $\x_i$ in cluster $j$ with $i \not = j^*$, the updates
\begin{align*}
    \hat{f}^*_t(\x_i) &= (1-\lambda)\frac{1}{k}\left((k-1)\hat{f}^*_{t-1}(\x_i) + \hat{f}^*_{t-1}(\x_{j^*})\right) + \lambda\hat{f}^*_{0}(\x_i)\\
    &= (1-\lambda)\frac{1}{k}\left((k-1)\hat{f}^*_{t-1}(\x_i) + \hat{f}^*_{t-1}(\x_{j^*})\right) + \frac{\lambda}{K}\mathbf{1}_K,
\end{align*}
whereas for the informative member of the cluster, we only have a different initialisation, and hence
\begin{align*}
    \hat{f}^*_t(\x_{j^*}) &= (1-\lambda)\frac{1}{k}\left((k-1)\hat{f}^*_{t-1}(\x_i) + \hat{f}^*_{t-1}(\x_{j^*})\right) + \lambda\hat{f}^*_{0}(\x_{j^*})\\
    &= (1-\lambda)\frac{1}{k}\left((k-1)\hat{f}^*_{t-1}(\x_i) + \hat{f}^*_{t-1}(\x_{j^*})\right) + \lambda\mathbf{e}_j,
\end{align*}
where $\mathbf{e}_j$ is equal to zero except in position $j$, where it takes the value one. It is then straightforward to show that the solution to which these converge is given by
\begin{align*}
    \lim_{t \to \infty} \hat{f}^*_t(\x_i) =& \frac{1-\lambda}{k}\mathbf{e}_j + \frac{k-1+\lambda}{kK}\mathbf{1}_K,\\
    \lim_{t \to \infty} \hat{f}^*_t(\x_{j^*}) =& \frac{1+\lambda(k-1)}{k}\mathbf{e}_j + (1-\lambda)\frac{k-1}{kK}\mathbf{1}_K.
%\\
%    \Rightarrow \frac{1}{n}\sum_{i=1}^n \max_{j\in [K]}(\ffinal)_{i,j} =& \frac{\lambda(K-1)}{n} + \frac{1-\lambda}{k} + \frac{k-1+\lambda}{kK}.%\\
    %=&: R(\lambda, k, K).%\\
    %&=\frac{k\lambda}{K(k-(1-\lambda)(k-\lambda))}\mathbf{1} + \frac{\lambda(1-\lambda)}{k-(1-\lambda)(k-\lambda)}\mathbf{e}_j.
\end{align*}
The improvement in clustering clarity under the idealised scenario can then easily be shown to be equal to
\begin{align*}
	\frac{1}{n}\sum_{i=1}^n &\left(\max_{j\in [K]}(\ffinal)_{i,j} - \max_{j\in [K]}(\finit)_{i,j}\right)\\
 & = \frac{\lambda(K-1)}{n} + \frac{1-\lambda}{k} + \frac{k-1+\lambda}{kK} - \frac{n-K+K^2}{nK}\\
 & = (1-\lambda)\left(\frac{1}{n} +\frac{1}{k} - \frac{1}{kK} - \frac{K}{n}\right).
\end{align*}
Including a reference value which depends on $k$ and $\lambda$ is important as it incorporates information relating to the flexibility of estimation. In particular, normalising $C(\lambda, k, K)$ by the improvement in clustering clarity under the idealised scenario will typically favour simpler models (with larger values of $k$ and $\lambda$), all other things being equal, and may be seen as penalising overly flexible models. However, without any prior information we do not want to favour any value for $K$ over others, and so we normalise $C(\lambda, k, K)$ by the largest potential improvement under the idealised scenario for the specified settings of $\lambda$ and $k$. That is, we compute
\begin{align*}
    R(\lambda, k)&:= \max_K \left\{(1-\lambda)\left(\frac{1}{n} +\frac{1}{k} - \frac{1}{kK} - \frac{K}{n}\right)\right\}\\
    &= (1-\lambda)\left(\frac{1}{n} + \frac{1}{k} - \frac{2}{\sqrt{nk}}\right),    
\end{align*}
and then perform selection from multiple models, for different settings of $\lambda, k, K$, by maximising $C(\lambda, k, K)/R(\lambda, k)$.
%it is also the case that $R(\lambda, k, K)$ is decreasing in $K$, the number of clusters, for small values of $K$. This may be undesirable since then normalising by this reference value will favour solutions with more clusters. To remove this effect, for each value of $k, \lambda$ we therefore normalise all corresponding values of $C(\lambda, k, K)$ by
%
%\begin{align*}
%    R(\lambda, k)&:= \min_K R(\lambda, k, K)\\
%    &= 2\sqrt{\frac{\lambda k (n-\lambda)}{n^2(k+1-\lambda)}} + \frac{(n-\lambda)(1-\lambda)}{n(k+1-\lambda)}.    
%\end{align*}
%
%That is, to select from among multiple potential solutions, for different values of $\lambda, k, K$, we select that which maximises $C(\lambda, k, K)/R(\lambda, k)$.

%\subsubsection{Large data sets}

%With the current implementation of our approach, combined estimation and selection on data sets containing up to about 10 000 points can be done efficiently, competitive in time to other non-parametric models such as spectral clustering (provided they also exploit the computational advantages of sparsity inducing weight/similarity functions). However, 

\section{Experiments}\label{sec:experiments}

In this section we present the results from experiments using 45 data sets taken from the public domain. All of these except for two\footnote{the Yale faces database~\url{http://cvc.cs.yale.edu/cvc/projects/yalefacesB/yalefacesB.html} and the Yeast data set, which was originally obtained from~\url{http://genome-www.stanford.edu/cellcycle/}, however the data set is no longer available at this address} were taken from the UCI Machine Learning Repository~\citep{UCI}. These are data sets for which the ground truth groupings of the data are known, and all have been used numerous times in the clustering literature, however as far as we are aware this is one of the largest collections of data sets used in any single study of this sort. Details of the data sets can be seen in Table~\ref{tb:metadata}, where the number of observations ($n$), dimensions ($d$) and true number of groups/clusters ($K$) are listed. Note that both the olive oil and frogs data sets have multiple potential ground truth label sets, and we report results for all of these. Before applying the different clustering methods we standardised all variables to have unit variance, after which we projected those data sets with more than 100 variables onto their first 100 principal components. Reducing dimensionality to a maximum of 100 was done purely for computational reasons, where the major computational bottleneck was associated with fitting large numbers of Gaussian mixture models on high dimensional data. To avoid the possibility that this dimensionality reduction either favours or disadvantages the GMM models, we simply applied all methods to the reduced data.
%
%For computational reasons we projected all datasets containing more than 100 variables onto their first 100 principal components. Importantly it is not the proposed approach whose running time on higher dimensional data is problematic, but rather it is the fitting of large numbers of Gaussian mixture models which presented the greatest computational burden.

\begin{table*}[]
    \centering
\scalebox{0.8}{
    \begin{tabular}{lrrr|lrrr|lrrr}
Data set & $n$ & $d$ & $K$ & Data set & $n$ & $d$ & $K$ & Data set & $n$ & $d$ & $K$\\
\hline
pendigits & 10992 & 16 & 10 & ionosphere & 351 & 33 & 2 & vowel & 990 & 10 & 11\\
optidigits & 5620 & 64 & 10 & banknote & 1372 & 4 & 2 & biodeg & 1055 & 41 & 2\\
mfdigits & 2000 & 216 & 10 & dermatology & 366 & 34 & 6 & ecoli & 336 & 7 & 8\\
wine & 178 & 13 & 3 & forest & 523 & 27 & 4 & led & 500 & 7 & 10\\
oliveoil & 572 & 8 & 3/9 & glass & 214 & 9 & 6 & letter & 20000 & 16 & 26\\
auto & 392 & 7 & 3 & heartdisease & 294 & 13 & 2 & sonar & 208 & 60 & 2\\
yeast & 698 & 72 & 5 & iris & 150 & 4 & 3 & vehicle & 846 & 18 & 4\\
yeast (UCI) & 1484 & 8 & 10 & libra & 360 & 90 & 15 & wdbc & 569 & 30 & 2\\
satellite & 6435 & 36 & 6 & parkinsons & 195 & 22 & 2 & wine & 1599 & 11 & 6\\
seeds & 210 & 7 & 3 & phoneme & 4509 & 256 & 5 & zoo & 101 & 16 & 7\\
imageseg & 2310 & 19 & 7 & votes & 434 & 16 & 2 & dna & 2000 & 180 & 3\\
mammography & 828 & 5 & 2 & frogs & 7195 & 22 & 4/8/10 & msplice & 3175 & 240 & 3\\
breastcancer & 699 & 9 & 2 & isolet & 6238 & 617 & 26 & musk & 6598 & 166 & 2\\
texture & 5500 & 40 & 11 & smartphone & 10929 & 561 & 12 & pima & 768 & 8 & 2\\
soybeans & 683 & 35 & 19 & yale & 5850 & 1200 & 10 & spambase & 4601 & 57 & 2\\

    \end{tabular}}
    \caption{List of data sets and their characteristics}
    \label{tb:metadata}
\end{table*}

\subsection{Clustering Methods}

A list of all clustering methods considered, and the approaches we used for model selection, is given below:
\begin{enumerate}
    \item $K$-means (KM): The classical clustering model, where we used the implementation in the {\tt R}~\citep{R} package {\tt ClusterR}~\citep{ClusterR}, and the popular $K$-means++ initialisation~\citep{ArthurV2007}. We used 10 initialisations due to the randomness in the $K$-means++ method, and selected the number of clusters using the silhouette score~\citep{kaufman2009finding}.
    %\item Ideal $K$-means: As above, except the ``best'' value for $K$, from the set $[30]$, is chosen as though the true groupings are available. This was done by taking $K$ to maximise the geometric mean of the Adjusted Rand Index~\citep{ARI} and Normalised Mutual Information~\citep{NMI}, computed on the solution with $K$ clusters and the true grouping. The number of existing approaches for selecting $K$ in a data driven way is extremely large, and so to avoid the difficult task of selecting from these merely for comparison purposes we include this idealised variant which represents an upper bound on the performance which could reasonably be expected from a data driven selection strategy.
    \item Gaussian Mixture Model (GMM): We used the implementation in the {\tt ClusterR} package, and considered a number of components up to $30$. We selected the number of clusters using the Bayesian Information Criterion~\citep{schwarz1978estimating}.
%    \item Spectral clustering (SC): We used a $k$NN affinity matrix with $k = \lceil 4 \log(n)\rceil$ as, after some experimentation with a range of settings, this was found to produce the most reliable results. We determined the number of (and allocation to) clusters using the approach described by~\cite{shefSC}, which uses a geometric argument that when the correct number of clusters (and hence eigen-vectors of the Laplacian) is identified they should align along distinct radii in the Laplacian eigen-space. Although the approach described by~\cite{shefSC} uses a dense affinity matrix, this could not be computed in reasonable time on the larger data sets considered. The geometric argument on which the method is based does not depend on the denseness of the affinity matrix, only that it has an embedded block-diagonal structure with blocks aligning with clusters. 
	\item Spectral Clustering (SC): We used the Spectral Partitioning Using Density Separation algorithm~\citep[SPUDS]{spuds}, as this was found to produce superior results to other {\tt R} implementations of SC of which we are aware. The method uses a Gaussian kernel similarity matrix and selects the number of clusters using a density separation criterion. We used the default settings described in the paper,
% and the implementation taken from \url{https://github.com/DavidHofmeyr/spuds}. In
 however in a few instances the method failed to produce an output. For these we simply incrementally increased the bandwidth parameter multiplier in the Gaussian kernel by 0.1 iteratively until a solution was obtained.% in the implementation in the in the {\tt R} package {\tt spuds}\footnote{taken from \url{https://github.com/DavidHofmeyr/spuds}}. Although this implementation provided better results overall than other {\tt R} implementations which we considered, it failed to produce a solution on some of the data sets and could not be run on the letters data set due to size.
    % which uses a $k$NN similarity matrix and automatically selects the number of clusters. We have found this to be one of the more accurate and computationally efficient implementations of spectral clustering available on the Comprehensive {\tt R} Archive Network (CRAN). %We fit models for $k \in \{5, 7, 9, 11, 13, 15\}$ and report the best performance from among these settings.
    \item HDBSCAN (HDB): The hierarchical variant~\citep{campello2013density} of the classical density based clustering algorithm~\citep{ester1996density}. The hierarchical model avoids the need to specify the bandwidth parameter in the density estimate. We set the number of neighbours required to classify a point as a ``high density point'' to each of $\{1/3, 1/2, 1, 2, 3, 4\}$ multiplied by $\log(n)$, and selected a solution using the Density Based Clustering Validation criterion~\citep[DBCV]{moulavi2014density}. DBSCAN and its variants do not allocate points not in the neighbourhood of a high density point to clusters, instead classifying them as noise. To make the results comparable with other methods, we merged these points with their nearest clusters. This was performed after selection using DBCV.
	\item Mean-shift (MS): We used a $k$NN mean-shift algorithm, and set $k = \lceil \log(n)\rceil$ as this was a setting which yielded the most consistently good results. We are not aware of any automatic approaches for selecting the number of neighbours for $k$NN mean-shift.
	\item Border Peeling Clustering (BPC): The approach proposed by~\cite{BPC}, which iteratively ``peels'' away points seen to be on the boundaries/borders of clusters, thereby exposing the cluster cores. It then uses a connectivity graph to partition the cluster cores, before assigning the peeled away points to one of these clusters based on information gathered during the peeling process. We used the code provided by the authors at~\url{https://github.com/
nadavbar/BorderPeelingClustering}, and the default settings recommended by the authors.
	\item Selective Nearest Neighbours Clustering (SNNC): The approach proposed by~\cite{SNNC} has a similar formulation to BPC, but uses fewer peeling iterations and different rules for defining the connectivity graph as well as for assigning border points to cluster cores. Again we used the code released by the authors, taken from~\url{https://github.com/
SSouhardya/SNNC}, and the default settings recommended by the authors.
	\item Torque Clustering (TC): The approach introduced
by~\cite{TC}, which produces a hierarchical clustering solution based on an agglomerative (bottom-up) algorithm. Clusters are iteratively merged with their nearest cluster of greater size (in terms of number of observations), and a final solution is obtained by cutting through the resulting tree-structure at the height corresponding to the largest gap in the merging statistics given by the product of the sizes of the two clusters being merged and the square of the distance between them. We used the implementation provided
by the authors at~\url{https://github.com/JieYangBruce/
TorqueClustering}.
    \item Clustering by Non-parametric Smoothing (CNS): The proposed approach. In preliminary experiments we found that setting $k$ on the order of $\log(n)$ and $\lambda$ on the order of $n^{-1/2}$ typically leads to reasonably good performance. We therefore fit models for $k \in \{\lfloor \log(n)\rfloor, 2\lfloor \log(n)\rfloor, 3\lfloor \log(n)\rfloor, 4\lfloor \log(n)\rfloor\}$ and $\lambda \in \{n^{-1/2}, 2n^{-1/2}, ..., 5n^{-1/2}\}$, and for $K$ up to 30, and select a model using the criterion described in the previous section. In a few cases the size of the set of candidates for the $\{\x_{i^*}\}_{i \in [K]}$, $K'$, was very large. We capped this number at 300, and when $K'$ initially exceeded 300 we simply included those with the 300 largest values of $||\W_{:i'}||_1 \min_{j \not = i} d(\x_{i'}, \x_{j'})$.

In addition, as we discuss below, we found that the Torque Clustering approach achieves very good performance especially when nearest neighbours are determined using the cosine distance. For a relevant comparison we also explore the use of the cosine distance within CNS, and denote the approaches using the regular Euclidean distance as TCe and CNSe, and the approaches using the cosine distance as TCc and CNSc.
\end{enumerate}

We also experimented with the Markov clustering algorithm, as implemented in the package {\tt MCL}~\citep{MCL}, however the running time for this method even on data sets of size 1000 was very substantial and we could not obtain solutions on the larger data sets. We also found the performance, when solutions could be obtained, to be poor in comparison with the other methods considered. We therefore do not include it in our comparison.

\subsection{Clustering Performance}

To assess the quality of clustering solutions obtained from the different methods, we compute the clustering ``Accuracy'', which is determined by the regular classification accuracy after an optimal permutation of the cluster labels; the Adjusted Rand Index~\citep[ARI]{rand1971objective}, which is the proportion of pairs of observations grouped either together or separately under both the clustering and true groups, adjusted by the expected proportion under random clustering; and the Adjusted Mutual Information~\citep[AMI]{AMI}, which is the mutual information in the clustering and true group partitions (again adjusted by the expectation under random allocation) and normalised to lie in the interval $[-1, 1]$.

\begin{table*}[h]
%\rotatebox[]{90}{
\centering
\scalebox{.6}{
    
    \begin{tabular}{l|rrrrrrrrrrrrrrrrrrrrrrrrrrrrrrrrrrrrrrrrrrrrrrrr}
          & 
\rotatebox{90}{pendigits} & 
\rotatebox{90}{optidigits} & 
\rotatebox{90}{mfdigits} & 
\rotatebox{90}{wine} & 
\rotatebox{90}{oliveoil ($C=3$)} & 
\rotatebox{90}{oliveoil ($C=9$)} & 
\rotatebox{90}{auto} & 
\rotatebox{90}{yeast} & 
\rotatebox{90}{yeast (UCI)} & 
\rotatebox{90}{satellite} & 
\rotatebox{90}{seeds} & 
\rotatebox{90}{imageseg} & 
\rotatebox{90}{mammography} & 
\rotatebox{90}{breastcancer} & 
\rotatebox{90}{texture} & 
\rotatebox{90}{soybeans}\\
\hline
GMM & 60.65 & 49.06 & 34.46 & 30.32 & 36.88 & 63.44 & 10.87 & 28.56 & 9.79 & 54.32 & 25.89 & 39.17 & 10.75 & 20.49 & {\bf 84.74} & 55.26\\
KM & 62.03 & 66.07 & 63.33 & {\bf 87.16} & 58.62 & 67.25 & 16.53 & 34.62 & 18.17 & 32.11 & 43.99 & 53.40 & {\bf 21.75} & 73.98 & 18.56 & 13.17\\
SC & 78.91 & 79.74 & 72.07 & 56.70 & 63.86 & 70.66 & 15.95 & 42.80 & 0.30 & 59.12 & {\bf 73.72} & 55.49 & 20.01 & 49.72 & 38.69 & 70.11\\
HDB & 72.40 & 0.30 & 0.30 & 0.79 & {\bf 79.62} & 55.23 & 24.46 & 0.63 & 2.58 & 14.62 & 46.50 & 40.99 & 11.61 & 45.90 & 24.34 & {\bf 70.53}\\
MS & 61.75 & 73.39 & 71.22 & 79.28 & 50.03 & {\bf 82.69} & 16.19 & {\bf 52.84} & 17.18 & 43.43 & 51.24 & 40.97 & 8.16 & 19.04 & 62.93 & 61.68\\
SNNC & 73.80 & 15.29 & 23.34 & -0.38 & 68.51 & 63.31 & 21.45 & 0.49 & 2.30 & 15.57 & -0.00 & 47.45 & 14.33 & 59.52 & 38.33 & 52.48\\
BPC & 73.84 & 0.70 & 81.53 & 0.00 & 76.14 & 59.50 & 24.34 & 0.00 & 0.00 & 51.95 & 42.03 & 50.20 & 20.37 & 76.96 & 64.70 & 50.86\\
TCe & 58.19 & 68.46 & 82.56 & 74.80 & 61.48 & 72.66 & 18.83 & 25.00 & 11.33 & 57.97 & 40.57 & {\bf 60.01} & 3.76 & 67.12 & 73.79 & 65.69\\
TCc & {\bf 81.57} & 78.80 & {\bf 84.01} & 72.30 & 71.71 & 62.73 & 24.98 & 19.53 & 9.29 & 20.68 & 61.28 & 59.20 & 21.63 & {\bf 78.45} & 32.61 & 68.97\\
CNSe & 81.00 & {\bf 80} & 82.63 & 40.23 & 61.43 & 68.85 & 18.60 & 46.44 & 2.58 & 55.00 & 69.40 & 52.33 & 12.00 & 29.42 & 72.35 & 40.98\\
CNSc & 78.61 & 78.26 & 79.48 & 81.91 & 58.17 & 74.05 & {\bf 25.07} & 48.44 & {\bf 21.53} & {\bf 65.21} & 65.90 & 49.88 & 9.32 & 34.47 & 65.32 & 68.68\\
\hline
&\rotatebox{90}{ionosphere} & 
\rotatebox{90}{banknote} & 
\rotatebox{90}{dermatology} & 
\rotatebox{90}{forest} & 
\rotatebox{90}{glass} & 
\rotatebox{90}{heartdisease} & 
\rotatebox{90}{iris} & 
\rotatebox{90}{libra} & 
\rotatebox{90}{parkinsons} & 
\rotatebox{90}{phoneme} & 
\rotatebox{90}{votes} & 
\rotatebox{90}{frogs ($C=4$)} & 
\rotatebox{90}{frogs ($C=8$)} & 
\rotatebox{90}{frogs ($C=10$)} & 
\rotatebox{90}{isolet} & 
\rotatebox{90}{smartphone}\\
\hline
GMM & 21.56 & 20.05 & 82.76 & 19.21 & {\bf 33.14} & 11.78 & 27.57 & 42.84 & 9.33 & 14.19 & 16.95 & 22.90 & 33.10 & 45.12 & 38.11 & 42.78\\
KM & 20.67 & 19.32 & 68.90 & 21.15 & 19.22 & 14.50 & 57.68 & 52.59 & 8.14 & 47.35 & {\bf 47.92} & 38.01 & 34.20 & 46.66 & 16.27 & 33.73\\
SC & 18.35 & 7.38 & 75.98 & 15.30 & 23.97 & 13.24 & 55.22 & 57.15 & 20.67 & 51.15 & 18.61 & {\bf 40.39} & {\bf 54.79} & {\bf 76.15} & 50.24 & 40.62\\
HDB & 16.84 & {\bf 43.79} & 29.11 & 16.13 & 15.50 & -0.43 & 57.68 & 54.72 & 8.95 & 49.89 & 47.40 & 38.20 & 54.50 & 73.35 & 1.27 & 0.08\\
MS & 18.58 & 17.54 & 80.54 & 32.04 & 25.55 & 13.55 & 52.05 & 46.73 & 12.75 & 80.08 & 14.09 & 23.51 & 33.36 & 44.36 & {\bf 71.75} & 48.93\\
SNNC & 14.57 & 39.89 & 64.68 & 2.10 & 25.75 & -0.21 & 57.39 & 20.74 & -1.03 & 0.86 & 27.63 & 17.64 & 15.97 & 12.93 & 5.70 & 47.20\\
BPC & 0.00 & 14.07 & 40.73 & 0.00 & 16.53 & 0.00 & 57.68 & 0.00 & 0.00 & {\bf 80.26} & 45.76 & 36.61 & 51.98 & 70.65 & 26.69 & 42.04\\
TCe & 11.94 & 21.17 & 78.64 & 24.97 & 21.38 & 10.74 & 57.68 & {\bf 58.8} & 9.39 & 79.12 & 23.51 & 34.43 & 51.78 & 61.03 & 56.50 & 41.64\\
TCc & 17.91 & 26.57 & {\bf 91.31} & 37.49 & 33.13 & 13.03 & 63.34 & 55.67 & {\bf 22.61} & 47.16 & 46.47 & 23.65 & 33.93 & 45.26 & 66.57 & 42.80\\
CNSe & {\bf 24.74} & 22.71 & 79.26 & {\bf 50.97} & 20.69 & {\bf 19.16} & 57.68 & 34.31 & 18.65 & 79.62 & 25.59 & 29.02 & 40.50 & 54.78 & 66.31 & 59.41\\
CNSc & 20.18 & 22.82 & 86.13 & 37.29 & 32.71 & 14.37 & {\bf 66.91} & 40.22 & 6.75 & 39.39 & 18.73 & 30.40 & 44.02 & 59.71 & 61.17 & {\bf 59.77}\\
\hline
&\rotatebox{90}{yale} & 
\rotatebox{90}{vowel} & 
\rotatebox{90}{biodeg} & 
\rotatebox{90}{ecoli} & 
\rotatebox{90}{led} & 
\rotatebox{90}{letter} & 
\rotatebox{90}{sonar} & 
\rotatebox{90}{vehicle} & 
\rotatebox{90}{wdbc} & 
\rotatebox{90}{wine} & 
\rotatebox{90}{zoo} & 
\rotatebox{90}{dna} & 
\rotatebox{90}{msplice} & 
\rotatebox{90}{musk} & 
\rotatebox{90}{pima} & 
\rotatebox{90}{spambase}\\
\hline
GMM & 79.87 & 39.92 & 9.06 & 42.84 & 42.57 & 39.73 & 0.00 & 21.32 & 39.59 & 5.71 & 73.83 & {\bf 14.98} & 0.00 & 3.42 & 2.26 & {\bf 11.81}\\
KM & 67.30 & 37.88 & {\bf 15.24} & 58.06 & 44.81 & 0.69 & 0.30 & 8.98 & 54.37 & 2.75 & 36.65 & 14.81 & 10.73 & 2.89 & {\bf 8.06} & 0.54\\
SC & 80.66 & {\bf 40.19} & 6.84 & 59.30 & 52.25 & 0.00 & 8.16 & 11.25 & 58.93 & 0.73 & 77.46 & 3.56 & 9.20 & 5.39 & 0.05 & 0.00\\
HDB & 14.03 & 28.80 & 1.16 & 5.56 & 7.94 & 37.99 & {\bf 10.03} & 0.37 & 2.80 & -0.00 & {\bf 86.61} & 0.52 & 5.68 & 3.06 & 0.51 & 0.13\\
MS & 51.50 & 39.33 & 6.77 & 46.45 & 44.41 & 48.48 & 9.27 & 15.58 & 42.33 & 7.62 & 65.98 & 4.21 & 4.64 & 5.06 & 2.01 & 8.88\\
SNNC & 40.34 & 0.00 & 2.44 & 38.65 & 53.33 & 0.00 & 4.74 & 3.52 & 0.38 & 0.31 & 72.62 & -0.17 & 0.26 & 1.66 & 1.14 & 5.14\\
BPC & 77.03 & 3.96 & 0.00 & 31.35 & {\bf 60.41} & 49.88 & 0.00 & 10.65 & 0.00 & 0.00 & 0.00 & 0.00 & 0.00 & {\bf 6.27} & 0.00 & 2.81\\
TCe & {\bf 86.32} & 37.48 & 7.14 & 36.91 & 38.92 & {\bf 54.85} & 6.81 & 16.15 & 38.06 & 4.98 & 76.22 & 7.26 & 4.91 & 4.90 & 3.60 & 6.97\\
TCc & 86.20 & 36.43 & 5.30 & {\bf 63.2} & 28.79 & 51.85 & 5.28 & {\bf 27.37} & {\bf 69.5} & 7.57 & 76.62 & 6.44 & 7.06 & 0.93 & 4.08 & 8.72\\
CNSe & 79.65 & 34.75 & 9.02 & 55.81 & 56.46 & 45.84 & 8.36 & 12.33 & 28.86 & {\bf 9.32} & 68.11 & 4.49 & {\bf 11.41} & 5.29 & 6.53 & 6.87\\
CNSc & 78.26 & 31.72 & 5.62 & 51.80 & 49.00 & 43.72 & 7.25 & 17.62 & 61.37 & 8.62 & 69.62 & 2.92 & 9.50 & 4.63 & 4.27 & 8.84\\
    \end{tabular}}%}
    \caption{Adjusted Mutual Information for all methods across all data sets}
    \label{tab:nmi_all}
\end{table*}

%\rotatebox{270}{
\begin{table*}[]
%\rotatebox[]{90}{
\centering
\scalebox{.6}{
    
    \begin{tabular}{l|rrrrrrrrrrrrrrrrrrrrrrrrrrrrrrrrrrrrrrrrrrrrrrrr}
 & 
\rotatebox{90}{pendigits} & 
\rotatebox{90}{optidigits} & 
\rotatebox{90}{mfdigits} & 
\rotatebox{90}{wine} & 
\rotatebox{90}{oliveoil ($C=3$)} & 
\rotatebox{90}{oliveoil ($C=9$)} & 
\rotatebox{90}{auto} & 
\rotatebox{90}{yeast} & 
\rotatebox{90}{yeast (UCI)} & 
\rotatebox{90}{satellite} & 
\rotatebox{90}{seeds} & 
\rotatebox{90}{imageseg} & 
\rotatebox{90}{mammography} & 
\rotatebox{90}{breastcancer} & 
\rotatebox{90}{texture} & 
\rotatebox{90}{soybeans}\\
\hline
GMM & 51.97 & 32.24 & 24.75 & 18.90 & 25.61 & 47.95 & 2.58 & 17.85 & 0.47 & 51.21 & 13.29 & 30.17 & 14.25 & 22.04 & {\bf 77.17} & 37.04\\
KM & 47.88 & 63.33 & 56.06 & {\bf 89.75} & 58.42 & 76.62 & -4.31 & 41.79 & 13.42 & 29.36 & 48.05 & 46.07 & 31.49 & 84.45 & 11.09 & 4.85\\
SC & 73.33 & {\bf 79.32} & 68.34 & 64.20 & 61.79 & 78.45 & 6.86 & 51.52 & -0.51 & {\bf 61.14} & {\bf 78.61} & 44.06 & 30.51 & 75.46 & 21.55 & 43.05\\
HDB & 59.65 & 0.00 & 0.01 & -0.52 & {\bf 89.81} & 52.88 & 5.41 & 0.46 & 1.16 & 7.83 & 48.83 & 22.46 & 12.49 & 74.13 & 8.54 & 41.73\\
MS & 50.46 & 70.29 & 67.31 & 80.01 & 51.67 & {\bf 85.44} & 12.37 & {\bf 53.55} & {\bf 14.99} & 37.82 & 53.33 & 19.99 & 3.24 & 18.10 & 46.75 & 36.47\\
SNNC & 62.52 & 4.74 & 9.32 & -0.82 & 79.18 & 60.16 & 4.67 & 0.14 & 1.33 & 8.16 & 0.00 & 24.84 & 28.68 & 82.54 & 14.45 & 35.91\\
BPC & 71.09 & 0.02 & 77.15 & 0.00 & 85.51 & 56.93 & 5.85 & 0.00 & 0.00 & 35.10 & 45.95 & 27.20 & 31.78 & 86.61 & 47.97 & 23.09\\
TCe & 36.94 & 57.22 & 76.71 & 74.14 & 62.53 & 77.42 & 14.63 & 29.50 & 11.05 & 59.75 & 45.26 & {\bf 51.38} & 0.28 & 78.59 & 61.87 & 41.92\\
TCc & {\bf 74.67} & 74.92 & {\bf 80.67} & 74.32 & 69.57 & 70.64 & 24.64 & 6.18 & 10.50 & 20.48 & 62.19 & 51.15 & {\bf 34.83} & {\bf 87.69} & 20.16 & 48.55\\
CNSe & 74.19 & 77.18 & 80.22 & 39.33 & 61.25 & 77.53 & 14.03 & 50.43 & 1.16 & 52.92 & 73.69 & 45.73 & 9.54 & 29.22 & 67.62 & 25.54\\
CNSc & 71.36 & 73.45 & 75.42 & 83.68 & 58.25 & 80.92 & {\bf 30.51} & 49.93 & 11.82 & 59.72 & 69.01 & 36.90 & 5.63 & 33.42 & 54.82 & {\bf 53.70}\\
\hline
&\rotatebox{90}{ionosphere} & 
\rotatebox{90}{banknote} & 
\rotatebox{90}{dermatology} & 
\rotatebox{90}{forest} & 
\rotatebox{90}{glass} & 
\rotatebox{90}{heartdisease} & 
\rotatebox{90}{iris} & 
\rotatebox{90}{libra} & 
\rotatebox{90}{parkinsons} & 
\rotatebox{90}{phoneme} & 
\rotatebox{90}{votes} & 
\rotatebox{90}{frogs ($C=4$)} & 
\rotatebox{90}{frogs ($C=8$)} & 
\rotatebox{90}{frogs ($C=10$)} & 
\rotatebox{90}{isolet} & 
\rotatebox{90}{smartphone}\\
\hline
GMM & 24.24 & 7.70 & 79.92 & 16.26 & 20.98 & 16.64 & 16.31 & 28.58 & 3.56 & 14.44 & 9.14 & 9.50 & 12.89 & 20.76 & 18.46 & 30.38\\
KM & 27.09 & 7.72 & 65.39 & 18.24 & 19.30 & {\bf 28.3} & 56.81 & 31.22 & -9.78 & 39.97 & 57.01 & 40.22 & 44.63 & 65.48 & 6.04 & 30.23\\
SC & 20.43 & 9.66 & 63.87 & 12.72 & 20.35 & 17.90 & 60.12 & 32.80 & {\bf 17.37} & 41.33 & 16.94 & {\bf 56} & 66.00 & 89.01 & 22.33 & 30.71\\
HDB & 18.66 & {\bf 55.73} & 20.67 & 12.94 & 13.77 & -1.04 & 56.81 & 32.98 & 1.65 & 41.63 & 56.32 & 54.61 & {\bf 68.44} & {\bf 92} & 0.07 & 0.01\\
MS & 18.23 & 5.39 & 78.40 & 17.14 & 11.78 & 17.47 & 49.32 & 28.05 & 8.44 & 70.41 & 8.66 & 10.57 & 14.30 & 22.60 & {\bf 48.21} & 42.76\\
SNNC & {\bf 33.38} & 52.49 & 55.86 & 1.91 & {\bf 25.77} & -1.61 & 55.84 & 8.21 & -2.63 & -0.21 & 46.18 & 15.46 & 14.38 & 9.62 & 0.41 & 31.76\\
BPC & 0.00 & 7.49 & 28.32 & 0.00 & 14.65 & 0.00 & 56.81 & 0.00 & 0.00 & {\bf 78.42} & {\bf 58.28} & 53.22 & 66.37 & 89.75 & 8.39 & 33.77\\
TCe & 19.69 & 9.84 & 73.16 & 16.07 & 12.40 & 11.68 & 56.81 & {\bf 35.81} & 1.92 & 77.30 & 24.25 & 47.28 & 56.79 & 78.16 & 32.31 & 22.41\\
TCc & 23.99 & 19.13 & {\bf 91.73} & 29.32 & 24.52 & 14.95 & 58.19 & 35.11 & 7.96 & 43.33 & 52.91 & 8.65 & 11.68 & 18.77 & 41.90 & 24.22\\
CNSe & 27.32 & 12.54 & 77.88 & {\bf 44.24} & 13.47 & 21.01 & 56.81 & 20.77 & 14.42 & 71.05 & 27.85 & 22.75 & 29.36 & 43.99 & 47.10 & {\bf 46.44}\\
CNSc & 24.13 & 15.72 & 86.37 & 28.60 & 21.54 & 17.42 & {\bf 62.74} & 25.52 & 3.30 & 32.13 & 17.28 & 28.43 & 37.02 & 54.24 & 38.90 & 45.31\\
\hline
&\rotatebox{90}{yale} & 
\rotatebox{90}{vowel} & 
\rotatebox{90}{biodeg} & 
\rotatebox{90}{ecoli} & 
\rotatebox{90}{led} & 
\rotatebox{90}{letter} & 
\rotatebox{90}{sonar} & 
\rotatebox{90}{vehicle} & 
\rotatebox{90}{wdbc} & 
\rotatebox{90}{wine} & 
\rotatebox{90}{zoo} & 
\rotatebox{90}{dna} & 
\rotatebox{90}{msplice} & 
\rotatebox{90}{musk} & 
\rotatebox{90}{pima} & 
\rotatebox{90}{spambase}\\
\hline
GMM & 69.84 & 19.37 & {\bf 8.7} & 41.22 & 35.98 & 12.94 & 0.00 & 15.22 & 55.75 & 3.09 & 83.87 & 7.67 & 0.00 & -0.77 & 1.07 & {\bf 10.86}\\
KM & 51.34 & 17.51 & 5.32 & 69.71 & 39.48 & 0.33 & -0.24 & 8.32 & 67.07 & 4.33 & 25.13 & {\bf 19.25} & {\bf 14.62} & 3.69 & {\bf 16.01} & -0.49\\
SC & 77.19 & {\bf 19.56} & 4.44 & {\bf 71.89} & 45.50 & 0.00 & 4.41 & 8.85 & 71.28 & 0.68 & 73.01 & 1.99 & 8.49 & {\bf 6.94} & 1.07 & 0.00\\
HDB & 3.48 & 7.09 & -2.35 & 3.80 & 4.09 & 9.41 & 3.75 & 0.06 & 2.77 & -0.22 & {\bf 95.12} & 0.42 & 0.81 & 0.03 & 1.46 & 0.12\\
MS & 26.69 & 18.59 & 3.80 & 36.80 & 37.52 & 16.44 & 5.46 & 7.68 & 48.08 & 5.31 & 50.93 & 4.09 & 3.98 & 0.25 & -0.40 & 7.01\\
SNNC & 8.63 & 0.00 & -5.42 & 43.22 & 48.40 & 0.00 & 0.31 & 0.36 & 0.91 & -0.54 & 69.09 & -0.40 & -0.34 & -5.87 & 3.29 & 1.54\\
BPC & 56.29 & 1.21 & 0.00 & 38.51 & {\bf 54.97} & 16.58 & 0.00 & 9.05 & 0.00 & 0.00 & 0.00 & 0.00 & 0.00 & 2.35 & 0.00 & 1.54\\
TCe & {\bf 77.79} & 17.63 & 4.07 & 42.19 & 33.45 & {\bf 27.49} & 4.63 & 10.31 & 43.02 & 1.38 & 70.01 & 12.01 & 1.15 & 0.43 & 1.89 & 3.82\\
TCc & 73.13 & 16.04 & 1.93 & 71.75 & 17.46 & 22.82 & 5.56 & {\bf 22.11} & {\bf 80.51} & 4.44 & 72.58 & 0.36 & 0.26 & 0.02 & 2.71 & 2.79\\
CNSe & 68.33 & 16.34 & 5.46 & 69.85 & 49.23 & 16.68 & {\bf 6.29} & 8.56 & 31.82 & 6.56 & 78.15 & 3.87 & 13.12 & 2.27 & 2.12 & 1.89\\
CNSc & 64.04 & 13.42 & 2.35 & 65.83 & 38.35 & 15.55 & 3.77 & 15.18 & 74.11 & {\bf 7.21} & 78.67 & 2.87 & 6.30 & 1.42 & 6.32 & 3.92\\

    \end{tabular}}%}
    \caption{Adjusted Rand Index for all methods across all data sets}
    \label{tab:ari_all}
\end{table*}

%\rotatebox{270}{
\begin{table*}[]
%\rotatebox[]{90}{
\centering
\scalebox{.6}{
    \begin{tabular}{l|rrrrrrrrrrrrrrrrrrrrrrrrrrrrrrrrrrrrrrrrrrrrrrrr}
 & 
\rotatebox{90}{pendigits} & 
\rotatebox{90}{optidigits} & 
\rotatebox{90}{mfdigits} & 
\rotatebox{90}{wine} & 
\rotatebox{90}{oliveoil ($C=3$)} & 
\rotatebox{90}{oliveoil ($C=9$)} & 
\rotatebox{90}{auto} & 
\rotatebox{90}{yeast} & 
\rotatebox{90}{yeast (UCI)} & 
\rotatebox{90}{satellite} & 
\rotatebox{90}{seeds} & 
\rotatebox{90}{imageseg} & 
\rotatebox{90}{mammography} & 
\rotatebox{90}{breastcancer} & 
\rotatebox{90}{texture} & 
\rotatebox{90}{soybeans}\\
\hline
GMM & 55.06 & 48.45 & 29.70 & 28.09 & 38.81 & 56.82 & 23.47 & 24.07 & 30.59 & 62.64 & 20.95 & 43.51 & 35.75 & 46.35 & {\bf 80.38} & 49.05\\
KM & 60.26 & 70.91 & 69.40 & {\bf 96.63} & 69.93 & 73.78 & 46.94 & 59.03 & {\bf 41.85} & 51.70 & 65.71 & 54.55 & {\bf 74.03} & 95.99 & 18.18 & 23.43\\
SC & 75.97 & 80.55 & 69.35 & 74.16 & 70.98 & 74.30 & 40.31 & 64.18 & 32.21 & {\bf 75.6} & {\bf 92.38} & 52.55 & 62.32 & 77.54 & 36.35 & {\bf 66.47}\\
HDB & 71.71 & 10.52 & 10.50 & 38.20 & {\bf 88.99} & 65.03 & 46.17 & 37.25 & 31.94 & 32.85 & 64.76 & 42.60 & 30.31 & 86.27 & 27.09 & 61.93\\
MS & 51.08 & 72.01 & 68.35 & 93.26 & 56.29 & {\bf 85.31} & 42.86 & {\bf 69.63} & 40.70 & 45.30 & 69.05 & 24.16 & 9.90 & 37.91 & 49.16 & 52.42\\
SNNC & 69.60 & 20.73 & 22.95 & 37.64 & 82.87 & 69.06 & 44.13 & 36.25 & 31.54 & 30.94 & 33.81 & 43.03 & 62.20 & 93.85 & 35.98 & 55.34\\
BPC & 72.78 & 10.94 & 83.60 & 39.89 & 86.19 & 67.31 & 45.15 & 36.53 & 31.20 & 52.17 & 64.76 & 46.41 & 62.92 & 96.57 & 56.51 & 47.29\\
TCe & 34.82 & 58.31 & 85.20 & 91.01 & 74.83 & 76.75 & 45.92 & 59.17 & 37.20 & 75.04 & 63.33 & {\bf 60.78} & 1.81 & 94.42 & 63.25 & 57.39\\
TCc & 81.96 & 79.61 & {\bf 91.05} & 91.01 & 82.52 & 64.86 & 63.27 & 14.61 & 33.15 & 42.50 & 84.29 & 57.23 & 67.03 & {\bf 96.85} & 27.27 & 57.69\\
CNSe & {\bf 82.94} & 80.30 & 88.35 & 60.11 & 71.33 & 74.13 & 46.68 & 64.04 & 31.94 & 63.51 & 90.48 & 49.87 & 24.28 & 49.93 & 69.75 & 40.12\\
CNSc & 80.33 & {\bf 84.18} & 85.10 & 94.38 & 65.73 & 79.37 & {\bf 66.33} & 64.18 & 36.99 & 74.08 & 88.10 & 43.72 & 14.61 & 52.65 & 64.93 & 61.93\\
\hline
 & 
\rotatebox{90}{ionosphere} & 
\rotatebox{90}{banknote} & 
\rotatebox{90}{dermatology} & 
\rotatebox{90}{forest} & 
\rotatebox{90}{glass} & 
\rotatebox{90}{heartdisease} & 
\rotatebox{90}{iris} & 
\rotatebox{90}{libra} & 
\rotatebox{90}{parkinsons} & 
\rotatebox{90}{phoneme} & 
\rotatebox{90}{votes} & 
\rotatebox{90}{frogs ($C=4$)} & 
\rotatebox{90}{frogs ($C=8$)} & 
\rotatebox{90}{frogs ($C=10$)} & 
\rotatebox{90}{isolet} & 
\rotatebox{90}{smartphone}\\
\hline
GMM & 50.14 & 12.46 & 84.15 & 44.74 & 49.07 & 44.22 & 25.33 & 37.22 & 20.00 & 38.99 & 18.66 & 20.71 & 27.37 & 37.73 & 19.06 & 41.72\\
KM & 65.53 & 12.17 & 71.86 & 50.67 & 47.20 & {\bf 71.09} & 66.67 & 40.56 & 60.00 & 58.55 & 87.79 & 68.49 & 63.17 & 68.69 & 7.69 & 33.80\\
SC & 53.85 & 65.89 & 69.40 & 47.61 & 42.99 & 53.40 & 74.00 & 46.11 & 50.26 & 59.99 & 36.18 & 69.49 & 71.81 & {\bf 82.71} & 32.82 & 38.37\\
HDB & 65.24 & {\bf 74.13} & 50.27 & 47.80 & 45.79 & 61.56 & 66.67 & 45.00 & 7.18 & 58.44 & 87.56 & 66.64 & {\bf 73.16} & 82.42 & 5.15 & 18.20\\
MS & 61.82 & 12.68 & 83.88 & 50.29 & 38.32 & 52.72 & 60.67 & 44.44 & 31.79 & 73.28 & 21.89 & 21.43 & 26.30 & 33.36 & {\bf 54.42} & 39.66\\
SNNC & 69.80 & 70.77 & 66.94 & 37.09 & 45.33 & 56.80 & 66.00 & 23.89 & 73.33 & 24.44 & 78.80 & 62.72 & 60.22 & 51.30 & 7.07 & 32.66\\
BPC & 64.10 & 29.88 & 55.74 & 37.28 & 46.26 & 63.95 & 66.67 & 6.67 & {\bf 75.38} & {\bf 83.41} & {\bf 88.25} & 61.13 & 66.75 & 76.82 & 13.53 & 37.00\\
TCe & {\bf 72.65} & 18.15 & 82.24 & 36.14 & {\bf 50.93} & 35.37 & 66.67 & {\bf 46.39} & 8.72 & 82.92 & 51.61 & {\bf 70.99} & 67.91 & 78.19 & 34.59 & 27.96\\
TCc & 47.58 & 28.21 & {\bf 95.36} & 57.36 & 46.26 & 54.08 & 79.33 & 45.83 & 65.13 & 55.25 & 86.41 & 19.05 & 26.34 & 36.25 & 47.77 & 30.93\\
CNSe & 67.52 & 17.06 & 83.33 & {\bf 74.76} & 40.19 & 61.56 & 66.67 & 30.56 & 69.23 & 73.87 & 50.92 & 38.37 & 44.32 & 54.93 & 49.81 & {\bf 50.52}\\
CNSc & 64.10 & 22.59 & 86.07 & 45.89 & 41.12 & 54.42 & {\bf 83.33} & 37.22 & 28.72 & 47.17 & 38.94 & 42.13 & 49.02 & 60.33 & 41.73 & 48.71\\
\hline
 & 
\rotatebox{90}{yale} & 
\rotatebox{90}{vowel} & 
\rotatebox{90}{biodeg} & 
\rotatebox{90}{ecoli} & 
\rotatebox{90}{led} & 
\rotatebox{90}{letter} & 
\rotatebox{90}{sonar} & 
\rotatebox{90}{vehicle} & 
\rotatebox{90}{wdbc} & 
\rotatebox{90}{wine} & 
\rotatebox{90}{zoo} & 
\rotatebox{90}{dna} & 
\rotatebox{90}{msplice} & 
\rotatebox{90}{musk} & 
\rotatebox{90}{pima} & 
\rotatebox{90}{spambase}\\
\hline
GMM & 69.11 & 28.08 & 59.15 & 64.88 & 51.00 & 30.07 & {\bf 53.37} & 34.87 & 78.21 & 18.01 & 86.14 & 18.90 & 51.91 & 17.20 & 16.41 & 32.91\\
KM & 54.17 & 24.04 & 62.75 & 77.08 & 55.00 & 5.71 & 52.40 & 36.88 & 91.04 & {\bf 46.72} & 39.60 & {\bf 58.8} & {\bf 55.02} & 41.83 & 52.86 & 59.86\\
SC & 76.97 & 29.39 & 48.15 & 77.38 & 66.00 & 4.06 & 21.15 & {\bf 38.42} & 92.27 & 42.71 & 82.18 & 15.70 & 38.96 & 38.38 & 64.58 & 60.60\\
HDB & 15.81 & 6.97 & 63.32 & 44.05 & 20.00 & 15.19 & 13.46 & 26.24 & 64.67 & 42.15 & {\bf 91.09} & 52.95 & 38.87 & 0.47 & 64.84 & 60.70\\
MS & 29.15 & 26.36 & 25.40 & 60.71 & 54.20 & 20.09 & 43.75 & 22.81 & 85.06 & 32.83 & 68.32 & 54.30 & 53.20 & 5.08 & 52.73 & 38.19\\
SNNC & 25.23 & 9.39 & 52.51 & 63.39 & 70.40 & 4.06 & 51.44 & 28.96 & 59.40 & 40.03 & 74.26 & 52.10 & 51.50 & {\bf 54.2} & 63.28 & 47.64\\
BPC & 61.25 & 14.14 & {\bf 66.26} & 62.50 & {\bf 76.00} & 28.73 & {\bf 53.37} & 37.83 & 62.74 & 42.59 & 40.59 & 52.55 & 51.91 & 26.43 & {\bf 65.1} & {\bf 61.66}\\
TCe & {\bf 85.35} & 30.71 & 38.86 & 63.69 & 48.60 & 32.09 & 30.77 & 23.88 & 83.30 & 8.13 & 75.25 & 32.05 & 14.20 & 5.84 & 15.89 & 16.28\\
TCc & 82.85 & 25.96 & 15.17 & {\bf 78.87} & 40.60 & 27.57 & 34.62 & 34.87 & {\bf 94.9} & 36.77 & 75.25 & 1.80 & 1.45 & 0.32 & 17.84 & 10.32\\
CNSe & 65.52 & {\bf 31.92} & 54.50 & 76.19 & 72.60 & {\bf 35.8} & 31.25 & 34.28 & 78.91 & 39.46 & 81.19 & 53.55 & 46.96 & 21.66 & 42.32 & 57.18\\
CNSc & 74.26 & 28.69 & 18.77 & 74.70 & 65.40 & 34.09 & 22.60 & 36.17 & 92.97 & 31.33 & 80.20 & 53.85 & 27.56 & 15.52 & 31.51 & 14.48\\

    \end{tabular}}%}
    \caption{Accuracy for all methods across all data sets}
    \label{tab:acc_all}
\end{table*}

\begin{table}[]
	\centering
    \scalebox{0.7}{
	\begin{tabular}{r|rrrrrrrrrrrr}
Performance Metric &	GMM & KM & SC & HDB & MS & SNNC & BPC & TCe & TCc & CNSe & CNSc\\
\hline
AMI & 6.83 & 6.09 & 5.16 & 7.11 & 5.94 & 8.39 & 7.48 & 5.36 & \textbf{4.38} & 4.64 & 4.62 \\
ARI & 6.92 & 5.94 & 4.82 & 7.19 & 6.44 & 8.16 & 7.09 & 5.50 & 4.73 & \textbf{4.57} & 4.65\\
Accuracy & 7.68 & 5.06 & \textbf{4.80} & 6.17 & 7.06 & 7.22 & 5.49 & 6.04 & 5.98 & 5.11 & 5.39 

	\end{tabular}}
\caption{Average rank of each method across all data sets, based on all three performance metrics}\label{tb:rank}
\end{table}

Tables~\ref{tab:nmi_all},~\ref{tab:ari_all} and~\ref{tab:acc_all} show the complete sets of results based on AMI, ARI and clustering accuracy, respectively. Note that we have multiplied the raw performance metrics by 100 to include more significant figures with fewer total digits, and in each case we highlight the highest performing method in bold font. Because of the very large number of individual results, we also provide some summaries of the contents of these in Tables~\ref{tb:rank},~\ref{tab:comp_nmi},~\ref{tab:comp_nmi}, and~\ref{tab:comp_nmi}. Specifically, in Table~\ref{tb:rank} we show the average rank of the performance of each method across all 48 clustering tasks, based on each of the metrics. In Table~\ref{tab:comp_nmi} we show the comparative performances between all pairs of methods, based on AMI. Specifically, we show the differences between the numbers of times the method listed row-wise outperformed the method listed column-wise and the converse. That is, if we let $\mathrm{AMI}_{ik}$ be the Adjusted Mutual Information of the $i$-th clustering method on the $k$-th data set, then the $i,j$-th entry in the table is given by
\begin{align*}
    \sum_{k=1}^{48} \left(I(\mathrm{AMI}_{ik} > \mathrm{AMI}_{jk}) - I(\mathrm{AMI}_{ik} < \mathrm{AMI}_{jk})\right).
\end{align*}
For example, in the row corresponding to CNSc and column corresponding to TCc, we see the value $-2$, meaning that TCc had higher AMI than CNSc two times more than CNSc had higher AMI than TCc. If there were no ties in performance, this would mean that TCc had higher AMI than CNSc 25 times out of the total 48. Tables~\ref{tab:comp_ari} and~\ref{tab:comp_acc} show the corresponding counts for ARI and clustering accuracy, respectively.

%To summarise the contents of these tables, we show in Tables~\ref{tab:comp_nmi},~\ref{tab:comp_ari},~\ref{tab:comp_acc} the comparative performances based on these three metrics, between all pairs of methods. Specifically, we show in the tables the differences between the numbers of times the method listed row-wise outperformed the method listed column-wise and the converse. That is, if we let $P_{ik}$ be the performance of the $i$-th clustering method on the $k$-th data set (using one of the metrics listed previously), then the $i,j$-th entry in a table is given by
%
%\begin{align*}
%    \sum_{k=1}^{48} \left(I(P_{ik} > P_{jk}) - I(P_{ik} < P_{jk})\right).
%\end{align*}
%
%For example, in Table~\ref{tab:comp_nmi} in the row corresponding to CNSc and column corresponding to TCc, we see the value $-2$, meaning that TCc had higher AMI than CNSc two times more than CNSc had higher AMI than TCc. If there were no ties in performance, this would mean that TCc had higher AMI than CNSc 25 times out of the total 48.

%\setlength{\tabcolsep}{4pt}

\begin{table*}[]
    \centering
    \scalebox{0.8}{
    \begin{tabular}{l|rrrrrrrrrrr}
   & GMM & KM & SC & HDB & MS & SNNC & BPC & TCe & TCc & CNSe & CNSc\\
   \hline
GMM & & -10 & -18 & 2 & -12 & 22 & 12 & -12 & -22 & -20 & -22\\
KM & 10 & & -12 & 7 & 2 & 24 & 9 & -9 & -8 & -20 & -12\\
SC & 18 & 12 & & 22 & 4 & 23 & 18 & -2 & -6 & -6 & -2\\
HDB & -2 & -7 & -22 & & -14 & 4 & 6 & -15 & -26 & -15 & -16\\
MS & 12 & -2 & -4 & 14 & & 24 & 10 & 2 & -12 & -16 & -22\\
SNNC & -22 & -24 & -23 & -4 & -24 & & -6 & -26 & -36 & -30 & -34\\
BPC & -12 & -9 & -18 & -6 & -10 & 6 & & -19 & -30 & -22 & -22\\
TCe & 12 & 9 & 2 & 15 & -2 & 26 & 19 & & -8 & -2 & -10\\
TCc & 22 & 8 & 6 & 26 & 12 & 36 & 30 & 8 & & 6 & 2\\
CNSe & 20 & 20 & 6 & 15 & 16 & 30 & 22 & 2 & -6 & & 6\\
CNSc & 22 & 12 & 2 & 16 & 22 & 34 & 22 & 10 & -2 & -6 & \\

    \end{tabular}}
    \caption{Pairwise comparative performances: Values in the table show the differences in the numbers of times the method listed row-wise outperformed the method listed column-wise and the converse, based on Adjusted Mutual Information}
    \label{tab:comp_nmi}
\end{table*}

\begin{table*}[]
    \centering
    \scalebox{0.8}{
    \begin{tabular}{l|rrrrrrrrrrr}
    &    GMM & KM & SC & HDB & MS & SNNC & BPC & TCe & TCc & CNSe & CNSc\\
    \hline
GMM & & -12 & -22 & 2 & -2 & 10 & 10 & -12 & -18 & -20 & -24\\
KM & 12 & & -16 & 11 & 4 & 20 & 1 & -1 & -10 & -11 & -4\\
SC & 22 & 16 & & 20 & 16 & 25 & 18 & 2 & -4 & -2 & 0\\
HDB & -2 & -11 & -20 & & -10 & 6 & -2 & -17 & -26 & -14 & -18\\
MS & 2 & -4 & -16 & 10 & & 16 & 4 & -6 & -4 & -24 & -20\\
SNNC & -10 & -20 & -25 & -6 & -16 & & -16 & -24 & -30 & -28 & -32\\
BPC & -10 & -1 & -18 & 2 & -4 & 16 & & -17 & -26 & -23 & -24\\
TCe & 12 & 1 & -2 & 17 & 6 & 24 & 17 & & -8 & -7 & -12\\
TCc & 18 & 10 & 4 & 26 & 4 & 30 & 26 & 8 & & -2 & -2\\
CNSe & 20 & 11 & 2 & 14 & 24 & 28 & 23 & 7 & 2 & & 6\\
CNSc & 24 & 4 & 0 & 18 & 20 & 32 & 24 & 12 & 2 & -6 &  \\

    \end{tabular}}
    \caption{Pairwise comparative performances: Values in the table show the differences in the numbers of times the method listed row-wise outperformed the method listed column-wise and the converse, based on Adjusted Rand Index}
    \label{tab:comp_ari}
\end{table*}

\begin{table*}[]
    \centering
    \scalebox{0.8}{
    \begin{tabular}{l|rrrrrrrrrrr}
& GMM & KM & SC & HDB & MS & SNNC & BPC & TCe & TCc & CNSe & CNSc\\
\hline
GMM & & -22 & -28 & -13 & -8 & -4 & -14 & -10 & -9 & -24 & -29\\
KM & 22 & & -6 & 19 & 18 & 26 & -1 & 3 & 2 & 1 & 6\\
SC & 28 & 6 & & 17 & 16 & 17 & 8 & 4 & 4 & 6 & 9\\
HDB & 13 & -19 & -17 & & 4 & 18 & -8 & 3 & -4 & -4 & -2\\
MS & 8 & -18 & -16 & -4 & & -6 & -14 & -4 & -8 & -22 & -18\\
SNNC & 4 & -26 & -17 & -18 & 6 & & -24 & -10 & -6 & -16 & -10\\
BPC & 14 & 1 & -8 & 8 & 14 & 24 & & -3 & 5 & -3 & -3\\
TCe & 10 & -3 & -4 & -3 & 4 & 10 & 3 & & -2 & -9 & -10\\
TCc & 9 & -2 & -4 & 4 & 8 & 6 & -5 & 2 & & -8 & -8\\
CNSe & 24 & -1 & -6 & 4 & 22 & 16 & 3 & 9 & 8 & & 6\\
CNSc & 29 & -6 & -9 & 2 & 18 & 10 & 3 & 10 & 8 & -6 &\\

    \end{tabular}}
    \caption{Pairwise comparative performances: Values in the table show the differences in the numbers of times the method listed row-wise outperformed the method listed column-wise and the converse, based on Clustering Accuracy}
    \label{tab:comp_acc}
\end{table*}

%Because of the very large number of individual results (48 clustering problems, after accounting for the multiple label sets for the frogs and olive oil data sets; and 11 clustering methods, after including both distance metrics in TC and CNS) we include the total set of outcomes in the appendix, and include only a summary here. Specifically, for each of the 48 clustering problems we determined three standardisations of each of the three sets of performance metrics: (i) the studentised performances (obtained by subtracting the mean and dividing by the standard deviation); (ii) the $[0, 1]$ mapped performances (obtained by applying the linear transformation for which the values exactly span the $[0, 1]$ interval); and (iii) the ranking of the performances (with lower rank being better performance). The average standardised performances based on Adjusted Mutual Information are shown in Table~\ref{tb:avg_nmi}, and those based on Adjusted Rand Index and Clustering Accuracy in Tables~\ref{tb:avg_ari} and~\ref{tb:avg_acc} respectively. In each case the best performing method is highlighted in bold font.

%Tables~\ref{tab:nmi_all},~\ref{tab:ari_all} and~\ref{tab:acc_all} show the complete sets of results based on AMI, ARI and clustering accuracy, respectively. Note that we have multiplied the raw performance metrics by 100 to include more significant figures with fewer total digits.

We summarise the main take-aways of the results below:%. Note that some of these comments are based on the overall set of performances, and are not apparent only from the averages:
\begin{enumerate}
    \item TCc has the lowest (best) rank on average when measured by AMI, while CNSe and SC have the best average ranks based on ARI and Accuracy, respectively. These are consistent with the pairwise comparisons, in that TCc outperformed every other method more often that the converse in terms of AMI, and the same holds for CNSe and SC when measured by ARI and clustering accuracy.
	\item Both variants of the proposed approach (using both distance metrics) are among the top few performing methods according to all performance metrics. Arguably the worst performance from CNS is when using the cosine distance (CNSc) and with performance measured by clustering accuracy.
	\item Overall the proposed approach with the Euclidean distance (CNSe) appears to be superior to the cosine distance variant, although the differences are not very substantial.% While CNSc has slightly higher average performance according to AMI and ARI, CNSe has better clustering accuracy and also a better average rank when using all three performance metrics.
	\item Torque Clustering achieves very good clustering performance, especially when using the cosine distance (TCc). In fact TCc achieves the best performance (across all three metrics and all 48 clustering problems) more often than any other approach, however its performance is more variable than CNS leading to slighlty lower performance on average when viewed across all metrics. In addition, both variants of TC perform relatively poorly in terms of clustering accuracy. It is interesting to note that the difference between the two distance metrics has a considerably more substantial influence on the performance of TC than it does on CNS.
	\item The other two more recently developed approaches, Border Peeling and Selective Nearest Neighbours Clustering, performed poorly overall, and in fact the latter was the worst performing of all methods considered. It is worth noting, however, that the settings recommended by the authors of the associated papers include a fixed value for $k$, the number of neighbours. It is possible that a setting which is at least dependent on the size of the data set would have rendered better results, however in the absence of an appropriate model selection criterion we did not want to deviate from the authors' recommendations.
	\item The SPUDS spectral clustering algorithm (SC) is arguably the third best performing approach overall, and the best performing of the more ``classical'' approaches. %It is also the best performing approach overall as measured by clustering accuracy. 
	\item Despite its simplicity, $K$-means achieved quite good performance overall, typically around the average performance in terms of AMI and ARI and performing very well in terms of clustering accuracy.
%	\item Mean Shift performed reasonably well overall, however this was based on a selction for the number of neighbours

%The proposed approach achieves the best performance overall, when considering all three performance metrics and all three standardisations. This is justified by the fact that both variants (using the different distance metrics) achieve at least third best 

%Although the more performant of the Torque Clustering varients achieves similar performance based on AMI and ARI, its performance in terms of clustering accuracy is far worse.
\end{enumerate}

\section{Conclusions}\label{sec:conclusions}

In this paper we introduced a novel formulation of the clustering problem, and phrased it as an estimation problem where the estimand is a function from the input space to the probability distributions of cluster membership. We proposed an implicit estimation procedure based on iterative non-parametric smoothing, and discussed its closed form solution. We also described an intuitive and fully data driven criterion which can be used to perform model selection for the proposed approach, which can select both the level of flexibility of estimation and also the number of clusters. 

In experiments using a very large number of data sets, we found that the proposed approach, which we termed ``Clustering by Nonparametric Smoothing'' (CNS), yields quite consistently good performance in comparison with relevant benchmarks. Although we have included a large number of individual data sets in our experiments, with the intention of assessing the general and broad applicability of the proposed approach, we acknowledge the potential for bias in the results owing to the fact that we experimented with this same collection of data sets in preliminary phases of development of the proposed approach. It is noteworthy that the recently proposed ``Torque Clustering'' approach also achieved very good performance on these data sets despite no such potential for bias.

%\bibliographystyle{spcustom}
%\bibliography{Bibliography-MM-MC}

\end{document}